\documentclass[a4paper,fleqn]{cas-sc}
\usepackage{hyperref}
\usepackage{cleveref}
\usepackage[numbers]{natbib}
\usepackage{subcaption}
\usepackage{tabularx}

\usepackage{cas-common}
\ExplSyntaxOn
\bool_gset_true:N \g_stm_nologo_bool
\ExplSyntaxOff

\def\tsc#1{\csdef{#1}{\textsc{\lowercase{#1}}\xspace}}
\tsc{WGM}
\tsc{QE}

\begin{document}
\let\WriteBookmarks\relax
\def\floatpagepagefraction{1}
\def\textpagefraction{.001}
\let\printorcid\relax 

\shorttitle{}    

\shortauthors{Mingxuan Sun et al.}

\title[mode = title]{VisAlgae 2023: A Dataset and Challenge for Algae Detection in Microscopy Images}  

\author[1]{Mingxuan Sun\textsuperscript{1,}}  
\author[2]{Juntao Jiang\textsuperscript{1,}}  
\author[3]{Zhiqiang Yang} 
\author[4]{Shenao Kong}  
\author[5]{Jiamin Qi}  
\author[5]{Jianru Shang}  
\author[5]{Shuangling Luo}  
\author[6]{Wanfa Sun} 
\author[7]{Tianyi Wang} 
\author[8]{Yanqi Wang} 
\author[9]{Qixuan Wang} 
\author[4]{Tingjian Dai}  
\author[10]{Tianxiang Chen} 
\author[11]{Jinming Zhang} 
\author[12]{Xuerui Zhang} 
\author[13]{Yuepeng He} 
\author[1]{Pengcheng Fu} 
\author[3]{Qiu Guan} 
\author[14]{Shizheng Zhou}  
\author[1]{Yanbo Yu}  
\author[4]{Qigui Jiang} 
\author[7]{Teng Zhou} 
\author[7]{Liuyong Shi} 
\author[4]{Hong Yan} 
\cormark[1]
\ead{yanhong@hainanu.edu.cn}

\address[1]{State Key Laboratory of Marine Resource Utilization in South China Sea, Hainan University, Haikou 570228, China} 
\address[2]{College of Control Science and Engineering, Zhejiang University, Hangzhou 310027, China} 
\address[3]{College of Computer Science and Technology College of Software, Zhejiang University of Technology, Hangzhou 310014, China} 
\address[4]{School of Computer Science and Technology, Hainan University, Haikou 570228, China} 
\address[5]{School of Art and Design, Guangdong University of Science and Technology, Dongguan 523083, China} 
\address[6]{School of Information and Electronic Engineering, Zhejiang University of Science and Technology, Hangzhou 310023, China} 
\address[7]{Mechanical and Electrical Engineering College, Hainan University, Haikou 570228, China} 
\address[8]{ChiYU Intelligence Technology (Suzhou) Ltd, Suzhou 215416, China} 
\address[9]{China Academy of Information and Communications Technology, Beijing 100083, China} 
\address[10]{College of Physics and Information Engineering, Fuzhou University, Fuzhou 350108, China} 
\address[11]{College of Computer and Information Technology, China Three Gorges University, Yichang 443002, China} 
\address[12]{College of Mathematics and Statistics, Chongqing university, Chongqing 401331, China} 
\address[13]{College of Computer Science, Chongqing university, Chongqing 401331, China} 
\address[14]{Institute of Applied Physics and Materials Engineering, University of Macau, Macau 999078, China} 

\cortext[1]{Corresponding authors at: School of Computer Science and Technology, Hainan University, Haikou 570228, China}  
\footnotetext[1]{These authors contributed equally to this work.}

\begin{abstract}
Microalgae, vital for ecological balance and economic sectors, present challenges in detection due to their diverse sizes and conditions. This paper summarizes the second "Vision Meets Algae" (VisAlgae 2023) Challenge, aiming to enhance high-throughput microalgae cell detection. The challenge, which attracted 369 participating teams, includes a dataset of 1000 images across six classes, featuring microalgae of varying sizes and distinct features. Participants faced tasks such as detecting small targets, handling motion blur, and complex backgrounds. The top 10 methods, outlined here, offer insights into overcoming these challenges and maximizing detection accuracy. This intersection of algae research and computer vision offers promise for ecological understanding and technological advancement.
\end{abstract}


\maketitle

\section{Introduction}
Microalgae, a remarkably diverse group of single-celled photosynthetic organisms, control many crucial aspects of global ecosystems. Their significance is not only deeply ingrained in the natural world but is also increasingly being recognized for its vast potential across a wide spectrum of applications, spanning environmental protection, ecological restoration, and energy production.

Microalgae possess a nutrient-rich chemical composition, which endows them with the status of being highly versatile resources. In the realm of food and animal feed, they can elevate the nutritional quality significantly. For instance, they are rich in proteins, vitamins, and essential fatty acids. These nutrients can enhance the growth and health of livestock and aquatic animals. In aquaculture practices, microalgae form the base of the food chain in aquatic ecosystems, providing a vital source of nutrition for fish larvae, mollusks, and crustaceans. Their presence in aquaculture ponds can improve the survival rate and growth performance of cultured organisms. Additionally, in the formulation of cosmetics~\cite{spolaore2006commercial}, microalgae extracts are being utilized for their unique properties, such as anti-aging, moisturizing, and antioxidant effects.

Meanwhile, microalgae are highly sensitive organisms, swiftly responding to even the slightest alterations in their surroundings. This characteristic makes them invaluable indicators for biomonitoring in both fresh waters~\cite{o2022microalgae} and oceans~\cite{desrosiers2013bioindicators}. Their widespread distribution across various aquatic habitats, from the smallest freshwater ponds to the vast expanse of the oceans, combined with their diverse taxonomy - there are thousands of different microalgae species - and rapid biomass accumulation, offers researchers a powerful tool. By studying microalgae populations and species composition, scientists can accurately assess water quality, ecosystem health, and the impacts of human activities on aquatic environments. 

Given the significance of microalgae as crucial environmental indicators, water sampling, when combined with microscopy imaging for algae analysis, offers a valuable perspective into environmental conditions. Traditional approaches to identifying and classifying algae species from microscope images demand substantial time and rely on highly trained professionals. This is precisely where the potential of AI-based computer vision technology, especially object detection, becomes prominent. Object detection plays multiple key roles as the joint tasks of classification and localization. In terms of classification, it can automatically recognize different algae species from microscope images, eliminating the need for manual, time-consuming identification. In terms of counting, object detection accurately tallies the number of algae, providing quantitative data for environmental assessment. By automating these processes, AI-based methods can process a large number of images rapidly, minimizing human error and significantly accelerating the speed of data analysis. This not only improves the efficiency of environmental monitoring but also ensures more accurate results, as stated in~\cite{zhou2023computer}.

This paper delves into presenting a dataset and outlines a challenge that took place in 2023. This challenge was an integral component of the IEEE Cybermatics 2023 conference. The dataset was carefully curated, containing a substantial collection of microscope images of various microalgae species. The challenge aimed to encourage researchers from around the world to develop innovative computer vision-based methods for accurate microalgae identification and classification. It further encompasses an in-depth overview of methods employed by participants who achieved Top 10 rankings on the challenge leaderboard. These methods ranged from advanced architecture design to preprocessing, augmentation, and post-processing methods, providing valuable hints to research in this field.


\section{Related Works}
\textbf{Object Detection}
Object detection is a core task in computer vision. Methods in the early stage are mainly based on feature extraction. For instance, Viola-Jones Detector~\cite{viola2001rapid} utilizes Haar features and  Histogram of Oriented Gradients (HOG)~\cite{dalal2005histograms} computes histograms of gradient orientations for each divided cell in images. In the past decade, the rise of deep learning~\cite{lecun2015deep} has driven significant progress in object detection. Deep learning-based detection methods can be divided into two categories. Two-stage methods first propose regions likely to contain objects, then classify those regions and refine their bounding boxes. Typical two-stage methods include RCNN~\cite{girshick2014rich}, Fast RCNN~\cite{girshick2015fast}, Faster RCNN~\cite{ren2015faster}, Cascade R-CNN~\cite{Cai2019} and so on. Single-stage methods directly detect objects without proposing regions, integrating classification and bounding box regression into a single step. Typical object single-stage methods include SSD~\cite{Liu2016} and YOLO series~\cite{redmon2016you,redmon2017yolo9000,redmon2018yolov3,bochkovskiy2020yolov4,yolov5,li2023yolov6,wang2022yolov7,yolov8_ultralytics,wang2024yolov9,THU-MIGyolov10,yolo11_ultralytics,tian2025yolov12}. Past years witnessed the success of Transformer in visual tasks. Transformer-based methods like DETR~\cite{detr}, Deformable DETR~\cite{zhu2021deformable} and RT-DETR~\cite{lv2023detrs} can reason about the relations of the objects and the global image context to enhance localizations. New techniques like the diffusion model have also been used in object detection. Diffusiondet~\cite{chen2022diffusiondet} formulates object detection as a denoising diffusion process from noisy boxes to object boxes, achieving competitive performance.

\textbf{Microalgae Detection} There have been some works utilizing classic or state-of-the-art methods for microalgae detection. Park et al. \cite{park2021microalgae} trained and evaluated the YOLOv3 model on a total of 1,114 algae images for 30 genera collected by microscope. Cao et al.\cite{cao2021detection} proposed an Improved YOLOv3 model for microalgae identification in ballast water, utilizing MobileNet as a lightweight backbone network, enhancing spatial pyramid pooling (SPP) for multi-scale feature extraction, and optimizing the loss function with the Complete IoU (CIoU). Liu et al.\cite{liu2022improved} proposed an enhanced Algae-YOLO object detection method utilizing ShuffleNetV2 as the backbone network to reduce parameters, integrating the ECA attention mechanism for improved accuracy, and employing ghost convolution modules in the neck structure for parameter size reduction. Yan et al.\cite{yan2023yolox} used an Improved YOLOx for multi-scale microalgae detection achieving high performance incorporating Focal and DIoU Loss, addressing the difficulty of imbalance inherent to microalgal detection. 

\textbf{"Vision Meets Algae" Series} "Vision Meets Algae" is a challenge series that focuses on developing algorithms for algae detection. The first "Vision Meets Algae"  challenge was held with IEEE UV2022 (the 6th IEEE International Conference on Universal Village) based on a dataset \cite{zhou2023vision} consisting of six genera of microalgae commonly found in the ocean (\textit{Pinnularia, Chlorella, Platymonas, Dunaliella salina, Isochrysis, and Symbiodinium}). The images of Symbiodinium in different physiological states known as normal, bleaching, and translating are also classified.


\section{The VisAlgae Challenge}
The goal of the VisAlgae 2023 challenge was to benchmark new and existing object detection algorithms for addressing the challenges of microalgal cell detection, focusing on the interdisciplinary application of algae research and computer vision technology. We conducted experiments on a high-throughput microfluidic platform, Collecting dynamic video data of microalgal cells under different fields of view and imaging conditions, followed by slicing the videos and carefully selecting frames to create the dataset. For specific details, please refer to our previous work \cite{zhou2024microfluidic,zhou2023vision}. Participants were provided access to the dataset, consisting of annotated training images and unannotated test images, and were asked to submit their results based on the test set. Participants need to overcome issues such as detecting small objects, handling multiscale issues, managing motion blur, dealing with complex backgrounds, and maximizing detection precision. 
\subsection{Data Description}
\subsubsection{Data Overview}
\begin{figure*}[htbp]
	\centering
    {\includegraphics[width=.95\linewidth]{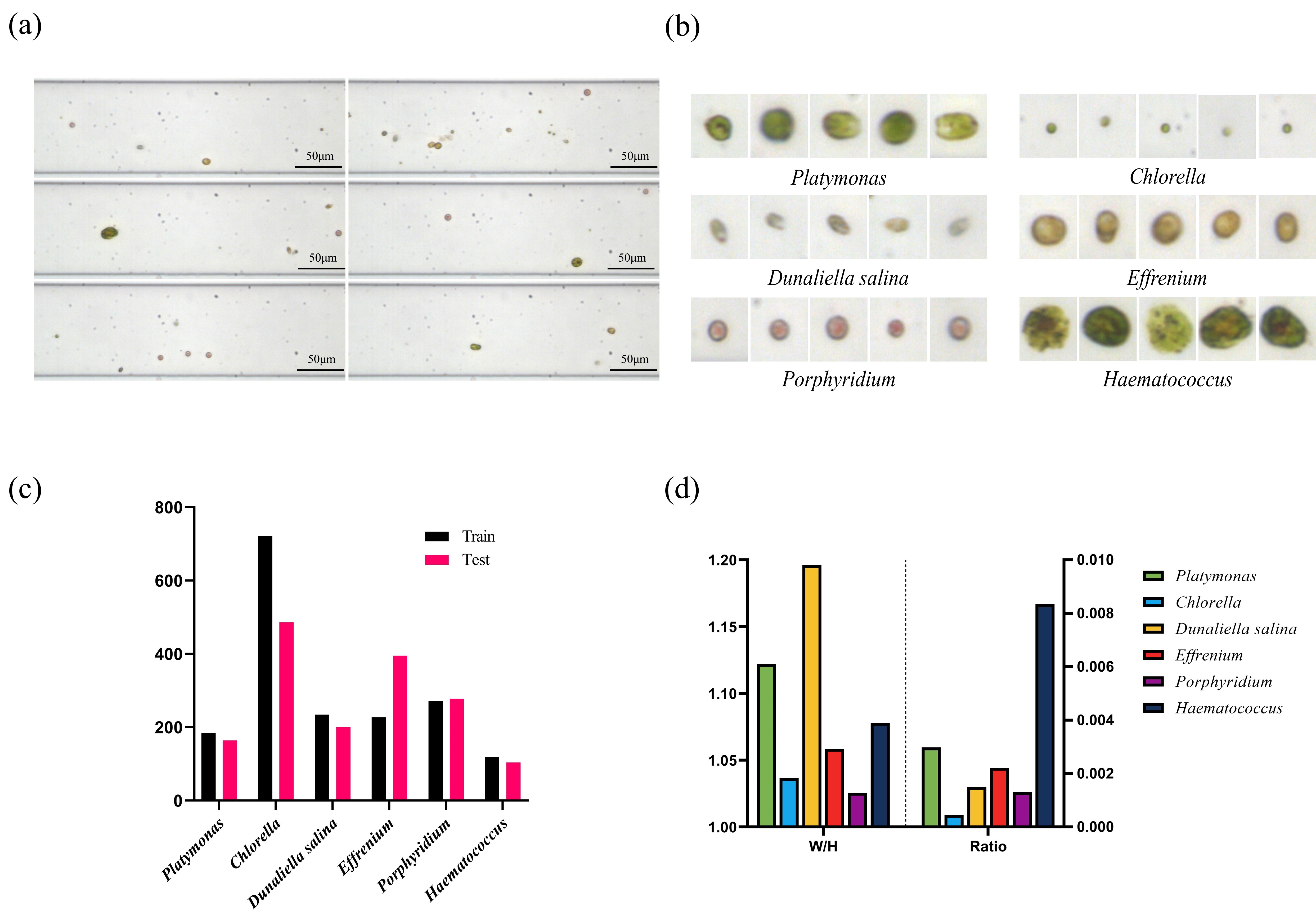}}\hspace{3pt}
	\caption{The statistics of annotations for each algal class are as follows: (a)Dataset images of algae in microfluidic channels; (b) object crops of each class; (c) the number of objects for each class; (d) the average aspect ratio and the ratio of their area to the entire image. "W/H" represents the aspect ratio, and "Ratio" represents the proportion of the area to the entire image.} 
    \label{Figure_1}
\end{figure*}

The complete VisAlgae 2023 dataset (training and testing) comprised 1000 images captured from six microalgae classes, as illustrated in Figure \ref{Figure_1}(a). The dataset includes six classes in total: \textit{Platymonas, Chlorella, Dunaliella salina, Effrenium, Porphyridium, and Haematococcus}, shown in Figure \ref{Figure_1}(b). The dataset is randomly scrambled and divided into a train set and a test set in a ratio of 7:3, and the two sets are independent of each other without duplicate images(train set: 700, test set: 300). The number of objects for each algal class in the train and test sets, along with their annotation box sizes, are depicted in Figure \ref{Figure_1}(c-d). It can be observed that there are a larger number of objects for \textit{Chlorella}, with annotation box sizes being particularly tiny. Additionally, the annotation box sizes for \textit{Haematococcus} and \textit{Chlorella} differ by approximately 5 times, highlighting the need for participant models to detect tiny objects while handling multi-scale scenarios. The color and morphological features of these six types of algae are shown in Table \ref{table1}. All images were obtained from the State Key Laboratory of Marine Resource Utilization in the South China Sea at Hainan University and were authenticated by domain experts. The annotations of the test set remained private to the challenge participants and accessible only to the challenge organizers, even during the evaluation phase.

\begin{table*}[h]
\centering
\caption{Summary of Six Algae Species (Simplified Features)}
\label{table1}
\begin{tabular}{>{\itshape}l l l}
\toprule
\textbf{Genus \&Species } & \textbf{Phylum} & \textbf{Key Features} \\
\midrule
Platymonas      & Chlorophyta         & Green; Flat oval \\
Chlorella       & Chlorophyta        & Green; Spherical\\
Dunaliella salina  & Chlorophyta      & Green; Oval or pear-shaped\\
Effrenium       & Dinophyta       & Yellow-brown;  Spindle-shaped\\
Porphyridium    & Rhodophyta           & Deep red-purple; Spherical \\
Haematococcus & Chlorophyta & Green to red; broadly ovate-elliptic\\
\bottomrule
\end{tabular}
\end{table*}

\subsubsection{Image Acquisition and Annotation}
The image dataset was acquired using an inverted microscopy platform (Olympus, IX73) with a connected industrial camera (MindVision Technology Co., Ltd, China). We captured images of six different algae mixtures at various resolutions to investigate their characteristics. Additionally, we collected cell images under different lighting and focusing conditions to introduce additional challenges. The amount of data and the quality of annotation are very important for neural network training. We manually annotated the images with LabelImg software. The annotations of the training set were then transferred to YOLO format. The annotation information for each image includes the positions, classes, and sizes of all the objects in the image. The object positions and sizes are represented by the center coordinates and dimensions respectively. As we expect this dataset to be used for other purposes to cross-modality domain adaptation, the data was released under a permissive copyright license (CC-BY-4.0), allowing for data to be shared, distributed, and improved upon. The detailed information of the dataset can be seen in Figure\ref{Figure_1}.
\subsection{Challenge Setup}
The testing phase was hosted on Alibaba Tianchi, a well-established competition platform that allows automated testing leaderboard management. Participant submissions are automatically evaluated using the pycocotools package. Each team can submit results up to five times a day, and the new results will automatically overwrite the old version. The testing phase was held between December 22, 2023, and January 25, 2024. 

\subsection{Metrics and Evaluation}
This challenge used mAP50:95 as the evaluation metric. Mean Average Precision (mAP) combines precision and recall values computed for each class and provides a single scalar value to represent the model's performance. The formula for calculating mAP is:
$$
m A P=\frac{1}{N} \sum_{i=1}^N A P_i,
$$

where $N$ is the total number of classes, $A P_i$ is the Average Precision computed for class $i$.

mAP50:95 refers to the mean Average Precision computed over a range of IoU thresholds from $50 \%$ to $95 \%$. This metric provides a comprehensive evaluation of the model's performance across various levels of IoU thresholds. 


\section{Baselines and Results}
For this dataset, we adopted various deep-learning object detection models with different sizes as baselines and conducted extensive experiments. Baseline models included YOLOv5s, YOLOv5mu, YOLOv5l~\cite{yolov5}, YOLOv8s, YOLOv8m, YOLOv8l~\cite{yolov8_ultralytics}, YOLOv9s, YOLOv9m, YOLOv9c~\cite{THU-MIGyolov10}, YOLOv10s, YOLOv10m, YOLOv10b~\cite{THU-MIGyolov10}, YOLOv11s, YOLOv11m and YOLOv11l~\cite{yolo11_ultralytics}, aiming to comprehensively evaluate the performance of each model in the algae detection task.
\subsection{Experimental Setup}
In the baseline experiments, we utilized an NVIDIA Tesla V100-16GB GPU for model training. The input image resolution was set to 640×640, and the training was conducted over 200 epochs with a batch size of 8. The initial learning rate was set to 0.01, and the final learning rate was set to 0.0001. To enhance the model's generalization capabilities, various data augmentation techniques were employed, including vertical and horizontal flipping, copy-paste augmentation, and rotation. Pretrained models on MS COCO are used. The dataset was split into training and validation sets with an 8:2 ratio.
\subsection{Experimental Results}
\begin{figure}[htbp]
	\centering
{\includegraphics[width=.95\linewidth]{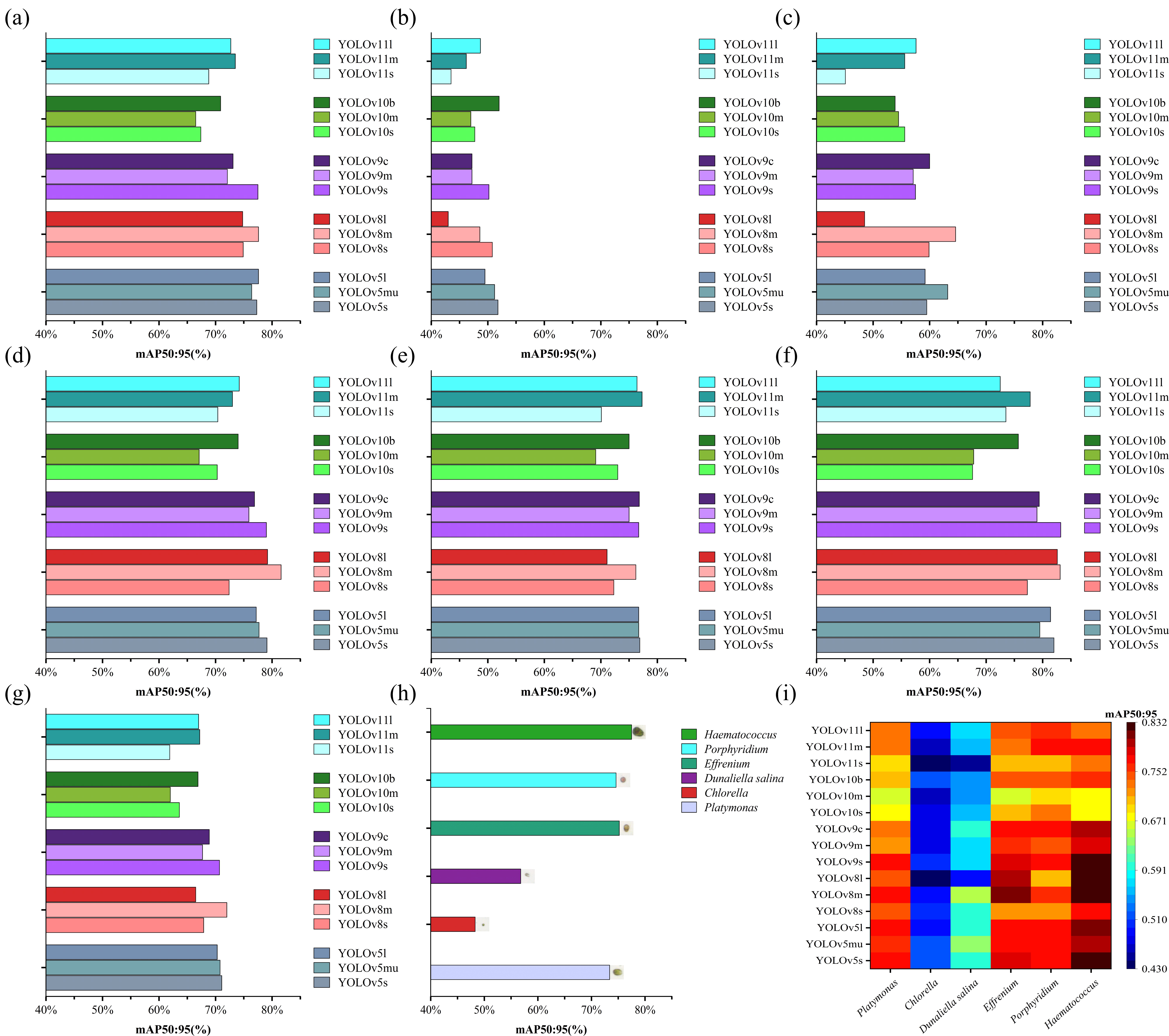}}\hspace{3pt}
\caption{The performance analysis of different baseline models in the algal classification task is as follows: (a)-(f) mAP50:95 on six algal classes: (\textit{Platymonas, Chlorella, Dunaliella salina, Effrenium, Porphyridium, Haematococcus})(g) Average performance across all classes. (h) Average per-class performance across all models. (i) Heatmap of mAP50:95 for each model and class.}
\label{Figure_2}
\end{figure}

The experimental results reveal different characteristics of object detection models between algal microscopy scenarios and natural images. Figure \ref{Figure_2}(a-f) shows the performance of these baseline models on six types of algae (\textit{Platymonas, Chlorella, Dunaliella salina, Effrenium, Porphyridium, Haematococcus}). In particular, the performance of the model does not correlate linearly with scale, illustrated by Figure \ref{Figure_2}(g), where the YOLOv5s demonstrates that the lightweight variant YOLOv5 achieved better performance (0.711) compared to its larger counterparts (YOLOv5mu/l: 0.69–0.70). This highlights that simply increasing model parameters fails to enhance feature representation for fine-grained algal morphology. The YOLOv8 series exhibited similar behavior, with YOLOv8m outperforming other variants (0.72 mAP50:95). As shown in Figure \ref{Figure_2}(h) and Figure \ref{Figure_2}(i), all models clearly exhibit poor performance on \textit{Chlorella}.

The underperformance of newer MS COCO-optimized models (e.g., YOLOv10, YOLOv11) stems from two factors:

\textbf{Domain Discrepancy:} Models that perform exceptionally well on MS COCO’s natural images often struggle when applied to microscopic algal features. This difficulty arises due to the distinct characteristics of microscopic imagery, such as translucent textures, low contrast, and fine-grained structures. Standard model architectures are typically designed to capture broad, generalizable patterns, which makes them less effective at preserving the domain-specific details crucial for accurate recognition in this specialized setting.

\textbf{Optimization Bias:}  While YOLOv10 prioritizes end-to-end efficiency through architectural innovations (spatial-channel decoupled downsampling, rank-guided block pruning) and YOLOv11 adopts multi-task modular designs (C3K2 dynamic kernels, partial self-attention), their shared emphasis on computational parsimony may inadvertently compress fine-grained spatial features \textemdash a critical limitation when differentiating algae species with subtle morphological variations. 

\section{Methods of Participants}
A total of 369 teams participated in the competition, with many achieving very high detection accuracy. The methodologies and results of the top-performing teams in the VisAlage Challenge are summarized in Table \ref{table2} and Table \ref{table3}. This section introduces the methods of the Top 10 teams, primarily focusing on data preprocessing and augmentation, architecture, inference optimization, and training strategies. 

\begin{table*}[htbp]
\centering
\caption{Summary of the ten top-performing methods in the VisAlgae Challenge.}
\label{table2}
\renewcommand{\arraystretch}{2} 
\resizebox{\textwidth}{!}{ 
\begin{tabular}{>{\raggedright\arraybackslash}p{2.1cm} 
                >{\raggedright\arraybackslash}p{3.6cm} 
                >{\raggedright\arraybackslash}p{5cm} 
                >{\raggedright\arraybackslash}p{4.2cm} 
                >{\raggedright\arraybackslash}p{2.8cm}}
\hline
\textbf{Team} & \textbf{Architecture} & \textbf{Preprocessing} \& \textbf{Augmentation} & \textbf{Post-training Optimization} & \textbf{Training Strategie}  \\
\hline
Z. Yang et al. & RTMDet-m with a neck using RepCSPLayer & Resize images to 1280$\times$1280. Apply Poisson Fusion, Copypaste, Mixup Copypaste, random scaling, flipping, rotation, and mosaic augmentation. Use a dynamic cache queue for faster target retrieval. Soft-NMS & TTA. Fuse predictions from four models using WBF. Remove slide noise using line detection. &\\ 
\hline
Y. Hu et al. & YOLOv5l, YOLOv8l, and Cascade R-CNN  & Resize images to varying resolutions (640$\times$640, 960$\times$960). Apply Poisson blending, Mixup, Mosaic augmentation, HSV transformation, translation, and random flipping.  & TTA. Fuse predictions from three models using WBF. & Multi-stage resolution fine-tuning.\\
\hline
W. Sun & Cascade R-CNN with backbones of ResNet-50, ResNet-101, and ResNeXt-101 & Resize input images to 1333$\times$800. Apply RandomFlip, RandomAffine, YOLOXHSVRandomAug, RandomShift, and JPEG compression. & Fuse predictions from eight models using WBF. &\\
\hline
S. Kong et al. & YOLOv5x with CBAM and Transformer modules & Apply Mixup and border box jitter.  &  &\\
\hline
Y. Wang & Cascade R-CNN with ResNeXt-101 backbone and FPN neck & Use original image resolution 1920×1200. Apply Mixup, Sharpen, and RandomFlip. Adjust receptive field size. &  & Utilize SWA for better generalization and optimize detection results. \\
\hline
Q. Wang & Cascade R-CNN and Co-DETR with multiple backbones (InternImage-L/XL, Swin-L, ConvNeXt V2, and FocalNet) & Resize images to either 11 fixed scales or high-resolution scales followed by random crops and resizing to lower resolutions. Apply random Copypaste. & Apply TTA with multi-scale inputs and horizontal flipping. Fuse predictions from ten models using WBF. & Multi-Scale Training\\
\hline
T. Dai et al. & YOLOv8x-p2  & Resize inputs to 2560$\times$2560. &  &\\
\hline
T. Chen & YOLOv5l with AIFI module and CARAFE upsampling & Resize images to 1280$\times$1280. Apply random scaling, cropping, panning, and rotation. Use MixUp, Mosaic augmentation, and noise addition to reduce background interference. &  &\\
\hline
J. Zhang & YOLOv8l and YOLOv6l-P6 & Resize images to random scales ranging from 0.5 to 1.5 times the original size. Apply image slicing and Copypaste. & TTA, Fuse predictions from two models using WBF. & Multi-Scale Training\\
\hline
X. Zhang et al. & YOLOv6-3.0 and YOLOv7-e6e & Resize images to 2560$\times$2560. Apply Poisson blending, random translation, rotation, noise, HSV transformation, and Mosaic augmentation. & TTA, Fuse predictions from two models using WBF. &\\
\hline
\end{tabular}
}
\end{table*}

\begin{table}[htbp]
\centering 
\caption{Final ranking of the VisAlgae challenge.}
\label{table3}
\begin{tabular}{p{1.5cm} p{3cm} p{1cm}}
\hline 
\textbf{Rank} & \textbf{Team}  & \textbf{Score} \\ 
\hline 
1 & Z. Yang et al. & 0.7604 \\
2 & Y. Hu et al. & 0.7477 \\
3 & W. Sun  & 0.7363 \\
4 & S. Kong et al. & 0.7360 \\
5 & Y. Wang & 0.7352 \\
6 & Q. Wang & 0.7340 \\
7 & T. Dai et al. & 0.7330 \\
8 & T. Chen & 0.7322 \\
9 & J. Zhang & 0.7292 \\
10 & X. Zhang et al. & 0.7244 \\
\hline 
\end{tabular}
\end{table}

\subsection{Preprocessing and Augmentation}
The competition entries in algae classification demonstrate both standardized and innovative approaches to data preprocessing and augmentation, offering valuable insights for algal detection. Key findings are synthesized as follows:

\subsubsection{Input Resolution}
Most teams maintained images in large resolutions (e.g., 1280$\times$1280, 960$\times$960, 1333$\times$800, 1920$\times$1200, 2560$\times$2560, and 3840$\times$2160) tailored to model specifications to enhance object detection performance. By maintaining higher pixel density, these large inputs preserve intricate visual details and contextual relationships, which are critical for resolving small objects, distinguishing fine-grained features, and minimizing background clutter interference. The increased spatial information in larger images provides more discriminative visual cues, enabling models to achieve superior localization accuracy. The comparison results of YOLOV7-e6e with different input sizes in the solution of the team in 10$^{\text{th}}$ place are shown in Table \ref{table4}. As shown in the table, larger input image sizes do not necessarily yield better results. The improvement in detection accuracy due to increased resolution is effective only within a certain range.

\begin{table}[htbp]
    \centering
    \caption{Comparison results of YOLOV7-e6e with different input sizes in the solution of the team in 10$^{\text{th}}$ place}
    \label{table4}
\begin{tabular}{p{3cm} p{1cm}} 
        \hline
        \textbf{Input Sizes} & \textbf{Score} \\
        \hline  
        1280$\times$1280 & 0.6076 \\
        2560$\times$2560 & 0.7027 \\
        3200$\times$3200 & 0.6976 \\
        \hline
    \end{tabular}
\end{table}

\subsubsection{Standard Augmentation Suite}
Geometric transformation techniques were widely used, including random scaling, flipping, rotation, random affine, random translation, and cropping, to increase image diversity. Color transformation methods like HSV transformation, JPEG compression corruption, and image sharpening were used to simulate different shooting conditions. As shown in Table \ref{Table5}, the 3$^{\text{rd}}$-place team presents the results of the ablation study on different data augmentation methods.

\begin{table}[htbp]
  \caption{Experimental results in the solution of the team in 3$^{\text{rd}}$ place}
\begin{tabular}{p{10cm} p{1cm}}
\hline \textbf{Methods} & \textbf{Score} \\
\hline baseline & 0.6671 \\
 baseline + RandomFlip & 0.7016 \\
baseline + RandomFlip + RandomAffine & 0.7032 \\
 baseline + RandomFlip  + YOLOXHSVRandomAug & 0.7051 \\
baseline + RandomFlip  + RandomShift & 0.7093 \\
 baseline + RandomFlip + RandomAffine + YOLOXHSVRandomAug & 0.6975 \\
 baseline + RandomFlip+ Corrupt(JPEG compression) & 0.7056 \\
\hline
\end{tabular}
\label{Table5}
\end{table}

\subsubsection{Advanced Innovations}

\begin{figure}
  \centering
  \begin{subfigure}[b]{0.3\textwidth}
    \includegraphics[width=\textwidth]{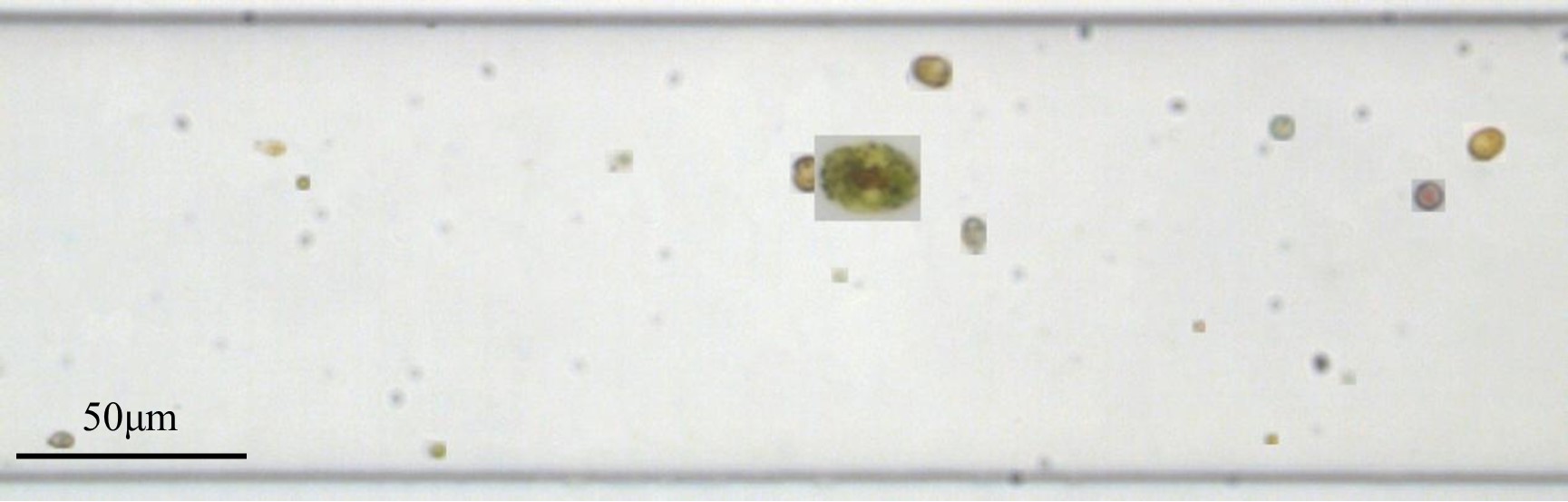}
    \caption{The augmented image after the CopyPaste method}
    \label{fig:sub1}
  \end{subfigure}
  \hfill
  \begin{subfigure}[b]{0.3\textwidth}
    \includegraphics[width=\textwidth]{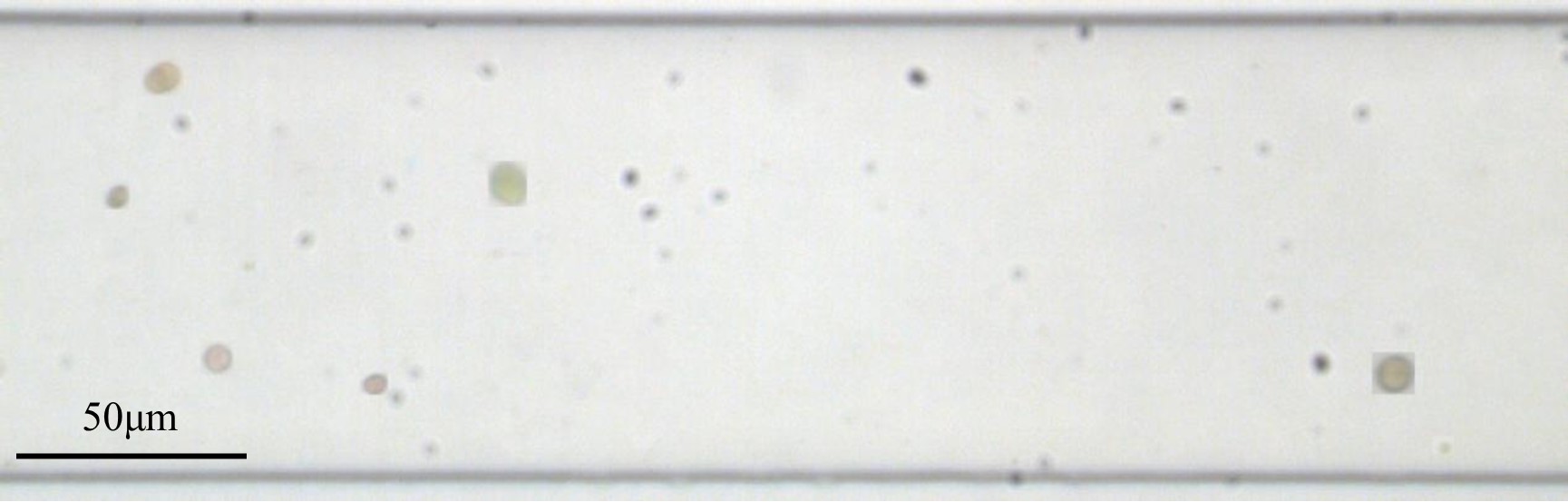}
    \caption{The augmented image after Mixup~\cite{zhang2018mixup}-Copypaste method}
    \label{fig:sub2}
  \end{subfigure}
  \hfill
  \begin{subfigure}[b]{0.3\textwidth}
    \includegraphics[width=\textwidth]{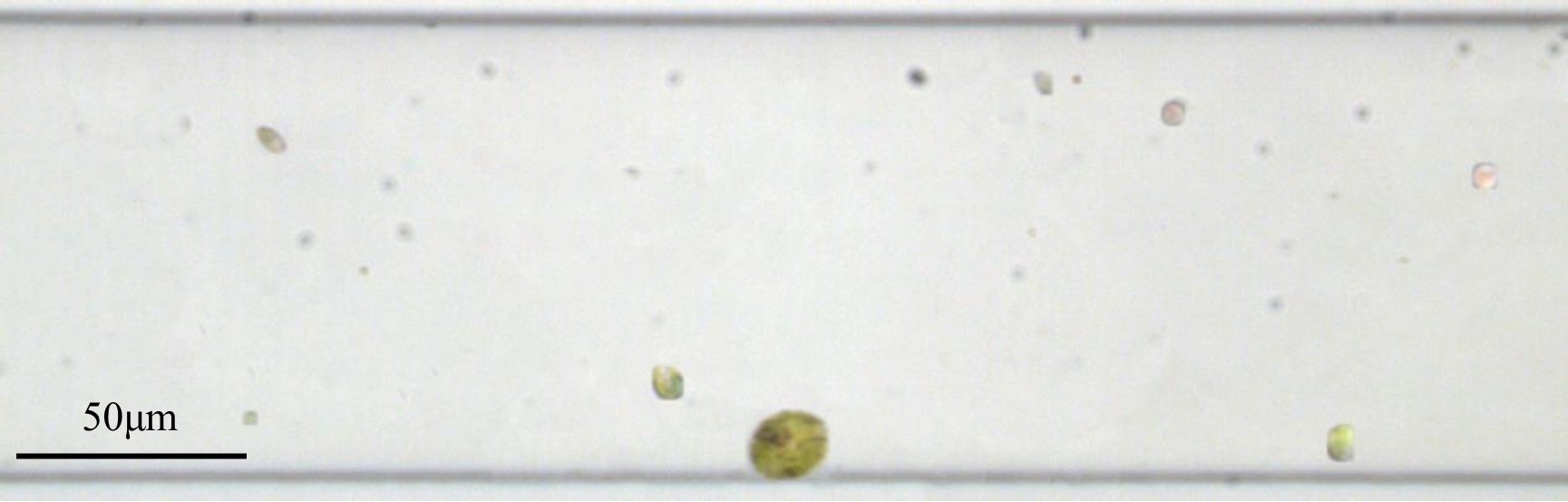}
    \caption{The augmented image after Possion fusion}
    \label{fig:sub3}
  \end{subfigure}
  \caption{Results of images after different augmentation methods from the solution of the team in solution of the team in 1$^{\text{st}}$ place}
  \label{Figure_3}
\end{figure}

\textbf{Fusion Operations:} Fusion operations such as Poisson Fusion, Copypaste, Mixup, and Mosaic are considered advanced methods, as shown in Figure \ref{Figure_3}. Poisson Fusion, for example, uses gradient-domain blending to seamlessly integrate foreground objects into different backgrounds, creating more realistic and diverse data samples. Copypaste~\cite{ghiasi2021simple} can transplant objects from one image to another, enriching the object distribution in the dataset. Mixup~\cite{zhang2017mixup} combines different images in a weighted-average manner, generating new samples that can help the model learn more complex feature relationships. The comparison of the effectiveness of these methods is shown in Table \ref{table6}.
Mosaic~\cite{bochkovskiy2020yolov4}, on the other hand, combines multiple images into one large image, increasing the context information and the complexity of the data. These operations are more complex than standard geometric and color transformations, and they can significantly expand the data diversity, thus having a more profound impact on improving the model's generalization ability.

The 1$^{\text{st}}$-place team introduced a dynamic cache queue mechanism into their Poisson blending augmentation pipeline. After each mosaic augmentation operation, algal targets from generated images are stored in the cache queue. The queue maintains a maximum capacity limit; once exceeded, older targets are randomly removed to preserve constant size. This mechanism accelerates target retrieval, thereby improving augmentation efficiency. Theoretically, the queue can cyclically store all algal targets across the dataset, enabling any training image to incorporate algal samples from all images via Poisson blending during augmentation. This design significantly enhances data stochasticity, expands training set diversity, and effectively mitigates overfitting risks.

The 1$^{\text{st}}$-place team also discussed the order of several augmentation methods and found that applying the dynamic cache-based Poisson fusion method augmentation method after mosaic augmentation yields better results, introducing greater randomness and allows for potential augmentation within the gray areas created by the mosaic.

\begin{table}[tb]
    \centering
    \caption{Comparative experiments of three different algae enrichment methods in the solution of the team in 1$^{\text{st}}$ place}
    \label{table6}
    \renewcommand\arraystretch{1.1} 
    \setlength{\tabcolsep}{8pt} 
    \normalsize 
    \begin{tabular}{c|c}
        \hline
        \textbf{Augmentation Methods} & \textbf{Score} \\
        \hline  
        - & 0.719 \\
        Copypaste & 0.721 \\
        Mixup-Copypaste & 0.726 \\
        Possion-Copypaste & 0.743 \\
        \hline
    \end{tabular}
\end{table}

\textbf{Noise Simulation for Complex Backgrounds:} Considering the practical challenges in real-world algae detection scenarios, including interfering factors such as water impurities or uneven lighting conditions, strategically injecting synthetic noise during training can simulate these environmental variabilities. By incorporating augmentation techniques like Gaussian noise injection, adaptive histogram distortion, and motion blur simulation, the model is forced to learn robust feature representations that disentangle algal morphology from transient artifacts, thereby improving both generalization performance and deployment reliability in heterogeneous aquatic environments. The 8$^{\text{th}}$-place team used this augmentation method.

\subsection{Architecture}
\subsubsection{Baseline Selection}
In the algae detection task, different teams have selected various architectures as baseline models, which are mainly divided into single-stage and two-stage detectors. While two-stage detectors may theoretically offer higher precision, single-stage models often achieve competitive or even superior results in the competition. This could stem from continuous architectural refinements in single-stage approaches that enhance their discriminative power, combined with the critical role of implementation. Additionally, their streamlined architectures may better adapt to small datasets, mitigating overfitting risks that could hinder more complex two-stage frameworks.

\textbf{Single-stage Detectors:} Most teams opted for single-stage detectors. Among them, the YOLO series is widely adopted, including YOLOv5l, YOLOv5x~\cite{yolov5}, YOLOv8l, YOLOv8x~\cite{yolov8_ultralytics}, YOLOv6-3.0~\cite{li2023yolov6}, and YOLOv7-e6e~\cite{wang2022yolov7}. Transformer-based method Co-DETR~\cite{zong2023detrs} is also utilized in this competition. Notably, the $1^{\text{st}}$-place team selected RTMDet-m~\cite{lyu2022rtmdet} as their baseline. RTMDet is a state-of-the-art real-time object detection framework that combines dynamic anchor-free detection, efficient feature pyramid networks, and novel loss functions to achieve a superior balance between speed and accuracy. Its optimized architecture enables high performance on both general and specialized datasets, making it a popular choice for real-world applications requiring low latency and high precision.

\textbf{Two-stage Detectors:} Teams that chose two-stage detectors all employed Cascade R-CNN~\cite{cai2018cascade} as their baseline due to the advantages of the cascade architecture, which includes multi-stage cascade training to progressively refine bounding box localization, adaptive IoU thresholds for hard-negative mining, and improved handling of objects across varying scales. Different backbones were adopted, including ResNet~\cite{he2016deep}, ResNeXt~\cite{xie2017aggregated}, InternImage~\cite{wang2023internimage}, Swin Transformer~\cite{liu2021swin}, ConvNeXt V2~\cite{woo2023convnext}, and FocalNet~\cite{yang2022focal}. Comparison results of Cascaded R-CNN and Co-DETR using different backbones in the solution of the team in 6$^{\text{th}}$ place are shown in Table \ref{table7}.

\begin{table}[htbp]
\caption{Test results of Cascaded R-CNN with different backbones in the solution of the team in 6$^{\text{th}}$ place} 
\label{table7}
\begin{tabular}{p{3cm} p{3cm} p{1.2cm}}
\hline 
\textbf{Models} & \textbf{Backbone} & \textbf{Score} \\
\hline 
\multirow{8}{*}{Cascade R-CNN}
  & InternImage-L & 0.6826 \\
 &  InternImage-XL & 0.7085 \\
  & Swin-B & 0.6653 \\
   & Swin-L & 0.6745 \\
  & ConvNeXt V2-B & 0.7206 \\
 & ConvNeXt V2-L & 0.7193 \\
  & FocalNet-B & 0.6904 \\
  & FocalNet-L & 0.6826 \\
  & FocalNet-XL & 0.6950 \\
  \hline 
 Co-DETR& Swin-L & 0.7265 \\
\hline 
\end{tabular}
\end{table}

\subsubsection{Improving Methods}
To improve the performance of the models, various enhancement modules have been introduced by different teams:

\textbf{Attention Mechanism Modules: }
To enhance feature representation, some teams incorporated attention-related modules. CBAM (Convolutional Block Attention Module)~\cite{woo2018cbam}, an attention mechanism, weights features in the channel and spatial dimensions. It focuses on local features, enabling the model to spotlight key algae features and suppress background noise. 

\begin{figure}
    \centering
    \includegraphics[width=0.6\linewidth]{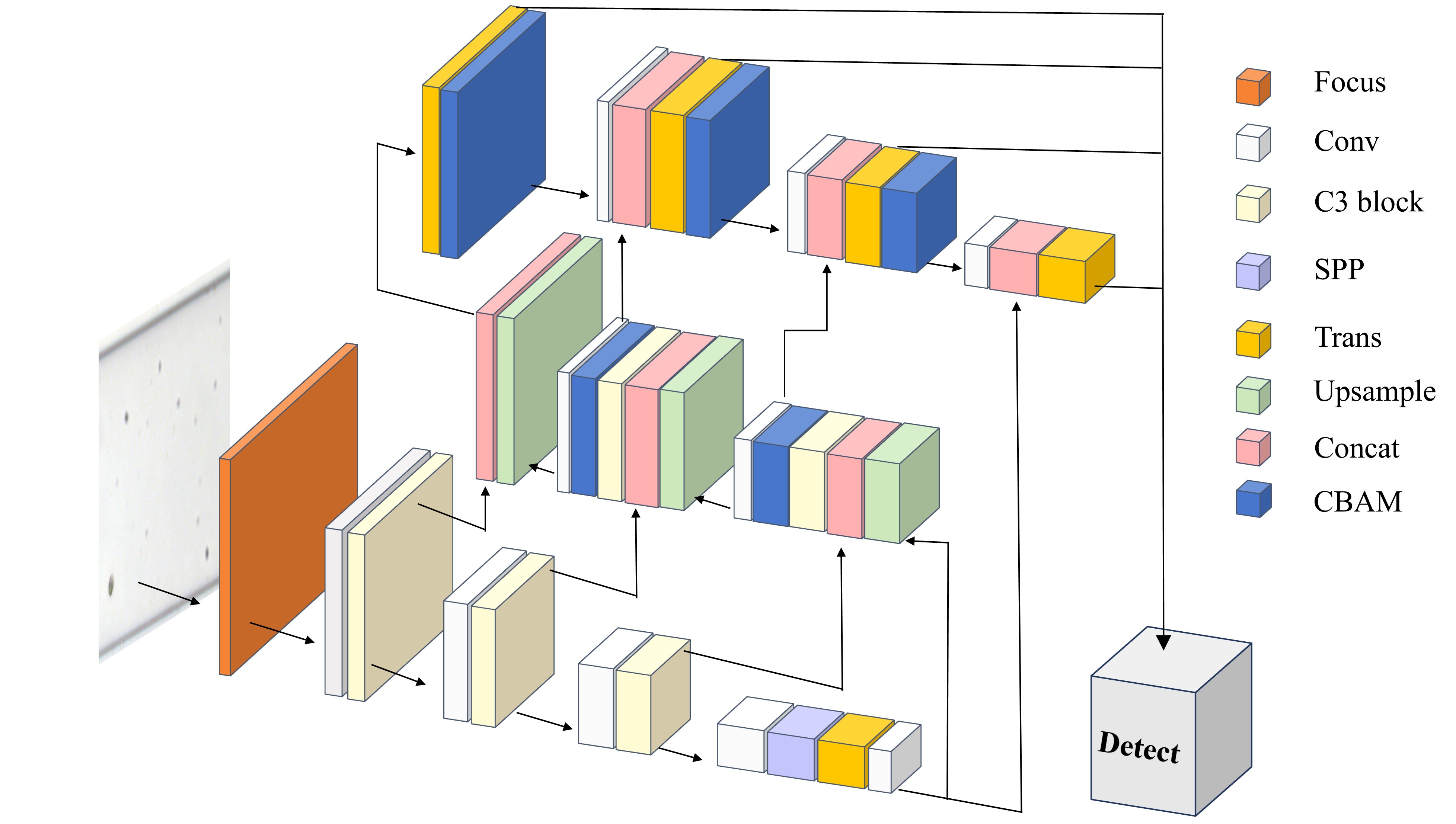}
    \caption{The architecture of the 4$^{\text{th}}$-place team improved YOLOv5 with CBAM and Transformer encoder blocks}
    \label{Figure_4}
\end{figure}

The transformer module~\cite{dosovitskiy2020image} can enhance the model's ability to capture long-range dependencies. They analyze the global context, helping the model better understand complex algae features in challenging backgrounds. They are more effective when dealing with targets with changeable morphologies or occluded. As shown in Figure \ref{Figure_4}, the $4^{\text{th}}$-place team used YOLOv5 as the baseline and incorporated both CBAM and Transformer encoder modules. 

\begin{figure}[tb]
    \centerline{
    \includegraphics[width=7cm]{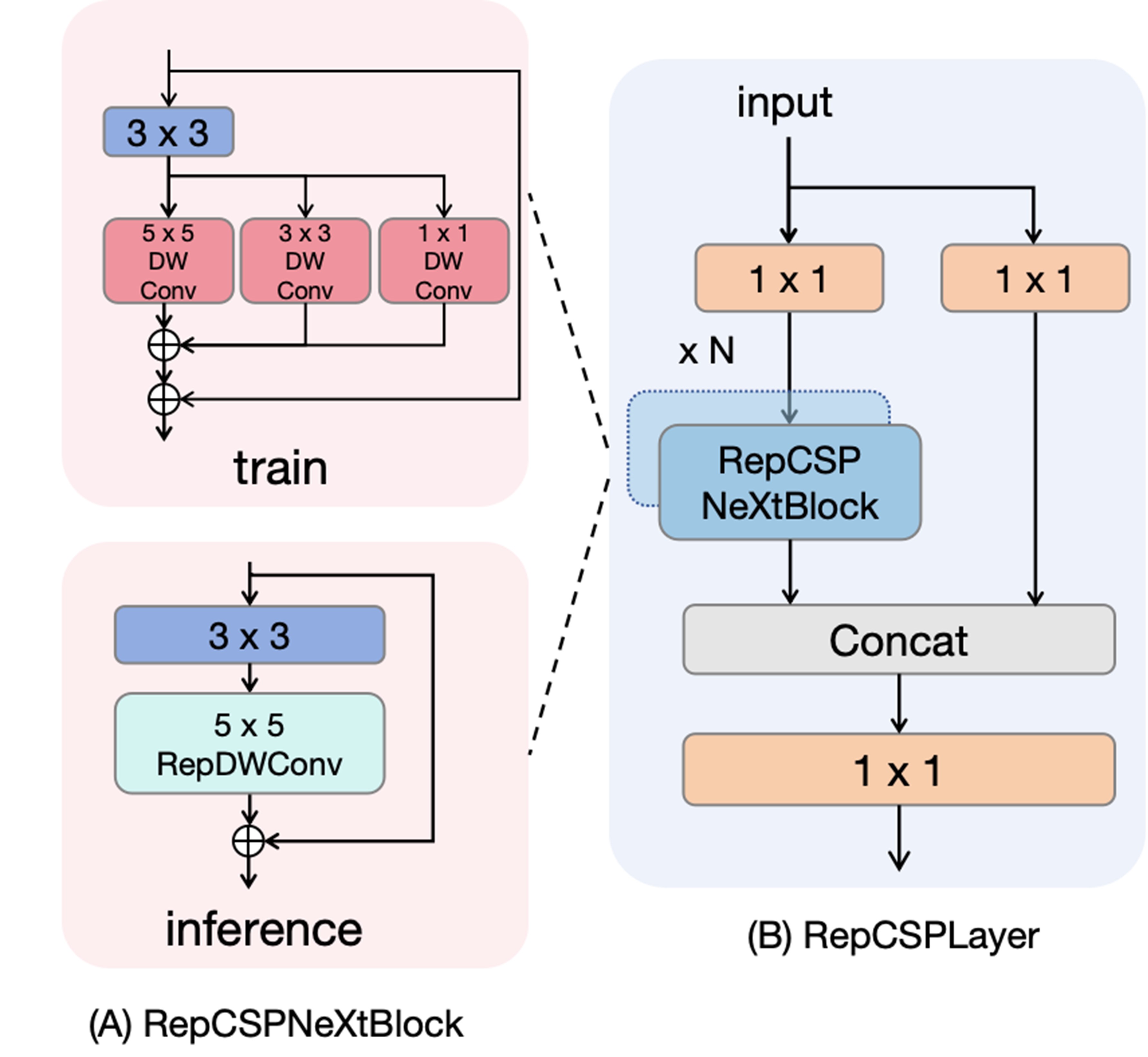}}
  \caption{The RepCSPLayer architecture in the solution of the team in 1$^{\text{st}}$ place}
  \label{Figure_5}
\end{figure}

\textbf{RepCSPLayer:}
The 1$^{\text{st}}$-place team proposed a RepCSPLayer structure to replace CSPLayer, employing the technique of reparameterized depthwise convolution. To maintain the stability of the shallow network structure, they only replaced all CSPLayers in the neck. The architecture of the RepCSPLayer can be seen in Figure \ref{Figure_5}. Through the combination of different branches, it enables the model to learn richer feature representations, thereby improving model performance.
Table \ref{table8} shows the performance of using RepCSPLayer in different positions and with depth-wise convolution of different kennel sizes, proving the effectiveness of the proposed method.

\begin{table}[htbp]
	\centering
	\caption{Comparative experiments of the 1$^{\text{st}}$-place team on RepCSPLayer structures.}
    \label{table8}
	\renewcommand\arraystretch{1}
	\setlength{\tabcolsep}{0.8mm}
		\begin{tabular}{p{3cm} p{2cm} p{2cm} p{2cm} p{1cm}}
		\hline
\textbf{Structure}& \textbf{Kernels} & \textbf{Location} & \textbf{RepDilated} &\textbf{Score}\\
		\hline
	  CSPLayer  & 5  &-& -&0.743 \\
		  RepCSPLayer  & 5, 3, 1&backbone& -&0.735\\
		  RepCSPLayer  & 5, 3, 1& neck&\checkmark&0.741\\
		RepCSPLayer  & 7, 5, 3, 1&neck& -&0.739 \\
	 RepCSPLayer  & 5, 3, 1&neck& -&0.7515\\

		\hline
	\end{tabular}
\end{table}

\textbf{Attention-based Intra-scale Feature Interaction (AIFI):} 
As the core component of RT-DETR~\cite{lv2023detrs}'s efficient hybrid encoder, the AIFI module applies single-scale Transformer encoders to the S5 feature layer for self-attention operations. This captures long-range dependencies between conceptual entities in high-level features, aiding subsequent modules in precise object localization and recognition. Comprising attention layers, feature fusion layers, and activation functions, the module uses dynamic weight allocation to enhance critical feature interactions. As shown in Table \ref{table9}, by replacing the SPPF in the backbone and neck of the YOLOv5l network with the AIFI module, The 8$^{\text{th}}$-place team achieved a 0.46\% improvement.

\textbf{CARAFE (Content-Aware ReAssembly of FEatures):} CARAFE~\cite{Wang_2019_ICCV} is a lightweight, content-aware upsampling operator. It dynamically generates instance-specific upsampling kernels via a lightweight convolutional module, enabling adaptive feature reassembly that preserves fine details and enhances semantic consistency. As shown in Table \ref{table9}, by integrating the CARAFE module into YOLOv5l with AIFI, The 8$^{\text{th}}$-place team achieved a 0.66\% improvement, representing a 1.12\% gain over the baseline YOLOv5l without enhancements.

\begin{table}[htbp]
\caption{Comparison results in the solution of the team in 8$^{\text{th}}$ place} 
\renewcommand\arraystretch{1}
\setlength{\tabcolsep}{2pt}
\label{table9}
\begin{tabular}{p{4cm} p{1cm}}
\hline
 \textbf{Methods} & 
\textbf{Score} \\
\hline
 YOLOv5l & 0.7210 \\
YOLOv5l+AIFI & 0.7256 \\
YOLOv5l+AIFI+CAREFE  & 0.7322 \\
\hline
\end{tabular}
\end{table}

\textbf{Connection Methods in YOLO:} The different connection methods of YOLO series models can also affect the model's attention to objects of different sizes. For example, the connection at P2 represents a feature map with a relatively high resolution in the YOLO model. It contains more fine-grained details, which makes it particularly suitable for detecting small objects. This is because small objects usually have fewer pixels and less obvious semantic features. Experiments of the 7$^{\text{th}}$-place team show that YOLOv8x achieved 0.718 while YOLOv8x-p2 achieved 0.733.

\subsection{Training Strategies}
\textbf{Multi-scale Training:} Multi-scale training is a technique in object detection where input images are dynamically resized during training to improve a model's ability to detect objects across different scales. By exposing the model to diverse image sizes, it learns robust multi-scale features, particularly enhancing performance on small objects. Methods include fixed-size random selection, dynamic range scaling with aspect ratio preservation, or batch-level adjustments for efficiency. This approach benefits small-object detection tasks, though it increases training time and memory usage. The team in 6$^{\text{th}}$ place and 9$^{\text{th}}$ place used this method.

\textbf{Stochastic Weight Averaging (SWA):}
SWA is a deep learning optimization technique that dynamically averages model parameters during training to enhance generalization. By periodically saving parameter snapshots (e.g., every 5 epochs) and averaging them at the end, SWA helps the model escape sharp local minima and converge to broader, flatter minima, improving robustness to unseen data. It often pairs with cyclic or fixed learning rate schedules to encourage parameter exploration. SWA is widely used in object detection tasks. The team in 5$^{\text{th}}$ place used this method.

\subsection{Inference Optimization}
The vast majority of teams adopted Weighted Box Fusion (WBF) and Test-Time Augmentation (TTA) as optimization techniques during the inference stage to enhance detection accuracy and robustness.

\textbf{Soft Non-Maximum Suppression (Soft NMS):} Soft NMS is a post-processing technique in object detection that improves on traditional NMS by reducing the confidence scores of overlapping bounding boxes rather than deleting them, preserving valuable information in dense scenes and avoiding missed detections caused by abrupt suppression. The 1$^{\text{st}}$-place team's experiments revealed that Soft NMS improves detection accuracy by 0.2\%.

\textbf{Weighted Box Fusion (WBF):}
WBF~\cite{solovyev2021weighted} is a post-processing technique in object detection that dynamically merges bounding boxes from multiple models or predictions. It groups overlapping boxes by IoU, calculates weighted-average coordinates based on confidence scores, and adjusts final confidence to retain complementary information from diverse sources. Unlike NMS, which suppresses overlapping boxes, WBF preserves all relevant predictions, improving detection accuracy in multi-model ensembles. Combinations of different models and their WBF fusion scores in the solution of the team in 6$^{\text{th}}$ place are shown in Table \ref{table10}. Fusion can be applied not only across different models but also across models trained on different folds. The 2$^{\text{nd}}$-place team employed a 5-fold cross-validation approach, then integrated the predictions of the five models using the WBF method, which is expected to enhance generalizability.

\begin{table}[htbp]
\caption{Combinations of different models and their WBF fusion scores in the solution of the team in 6$^{\text{th}}$ place} 
\label{table10}
\centering
\setlength{\tabcolsep}{3pt} 
\renewcommand{\arraystretch}{1} 
\begin{tabularx}{\textwidth}{p{15cm} p{1cm}} 
\hline
\textbf{Combination of different models (and models with different backbones)} & \textbf{Score} \\
\hline
Cascaded R-CNN (InternImage-L/-XL) & 0.7118 \\
Cascaded R-CNN (InternImage-L/-XL + Swin-B/-L) & 0.7294 \\
Cascaded R-CNN (InternImage-L/-XL + Swin-B/-L + ConvNeXt V2-B/-L) & 0.7311 \\
Cascaded R-CNN (InternImage-L/-XL + Swin-B/-L + ConvNeXt V2-B/-L + FocalNet-B/-L/-XL) & 0.7334 \\
Cascaded R-CNN (InternImage-L/-XL + Swin-B/-L + ConvNeXt V2-B/-L + FocalNet-B/-L/-XL) + Co-DETR (Swin-L) & 0.7340 \\
\hline
\end{tabularx}
\end{table}

\textbf{Test-Time Augmentation (TTA):}
TTA in object detection involves applying multiple data augmentations (e.g., flipping, scaling, rotating) to input images during inference, generating diverse predictions that are then aggregated to improve final results. Many participants have adopted TTA to boost their performance. For each augmented image, bounding boxes and class scores are adjusted to match the original image's coordinate system, and predictions are averaged or weighted to produce a refined detection. 

\textbf{Removal of Slide Noise}
The team in 1$^{\text{st}}$ place analyzed that the slide noise stems from small black dots and horizontal lines. Their test set background image analysis revealed no algae/impurities outside black lines, suggesting algae absence in these regions. To mitigate false positives from impurity misclassifications, they removed all line-external predictions. Leveraging consistent image perspectives, a Hough transform-based line detector localized boundaries across the dataset. Integrated into the WBF model, this denoising strategy slightly improved the score from 0.7593 to 0.7604.
\subsection{Frameworks}
 Many participants use detection frameworks like ultralytics~\cite{yolov8_ultralytics} and mmdetction~\cite{mmdetection}, hereby streamlining the development process and enabling the rapid iteration of innovative methods, which offers great convenience for them to develop their methods.

\subsection{Failed Attempts}
In the competition, some tricks were applied but failed to improve detection results. However, they yield valuable insights by uncovering hidden patterns, exposing limitations, or inspiring alternative strategies. These "failed" experiments refine approaches and foster critical thinking, advancing understanding even without measurable success.

In the algae detection task, a sliding window strategy was explored by the 1$^{\text{st}}$-place team to address small object challenges. Initial attempts using traditional sliding window and NMS fusion underperformed, prompting the team to make modifications including edge box removal, window padding, and WBF fusion to mitigate performance decline. Despite these adjustments, no significant improvements were achieved compared to non-sliding window methods. Additionally, various other strategies were tested by the 1$^{\text{st}}$-place team, such as modifying the label assignment cost matrix in RTMDet; incorporating multi-scale transformation factors into the loss function to increase the weight of larger targets; introducing a Distance-Focal Loss (DFL) localization head and loss; refining the FPN structure for richer fusion; merging backbone layers into the neck architecture; training with larger scales such as 1920 and 2560; further modifying the structure of the RepCSPLayer; and employing knowledge distillation to train smaller-scale models. While these approaches demonstrated theoretical potential, none yielded measurable improvements in detection performance. These iterative experiments highlight the complexity of balancing model design with dataset-specific characteristics.


\section{Discussion}
While recent YOLO generations achieve superior performance on natural images, state-of-the-art models trained on MS COCO may not outperform older architectures on algal microscopy images. This highlights the critical challenge of domain shift, where model generalization is hindered by differences in image characteristics (e.g., low contrast, dense clustering, and small object size) between natural and microscopic datasets. It underscores the importance of dataset-specific fine-tuning rather than blind reliance on generic pre-trained models.

 Generative methods have been explored for data augmentation in object detection, demonstrating effectiveness in improving model generalization and addressing challenges like class imbalance~\cite{trabucco2023effective, fang2024data, zhang2024enhancing}. However, in this competition, while basic augmentations like rotation are common, generating synthetic data from existing datasets remains underexplored. These methods can address issues like overfitting and domain shift by expanding training diversity, though validation is needed to ensure data integrity. As generative tools improve, systematic exploration of this strategy may unlock competitive advantages.

Post-processing techniques like WBF and TTA remain highly effective for accuracy gains in competitions, yet they are often impractical for real-world deployment due to real-time constraints. This discrepancy reveals limitations in competition metric design, which prioritizes pure detection performance over computational efficiency and latency. A more balanced evaluation framework incorporating both accuracy and inference speed would better reflect practical requirements.

Strategies tailored to small object detection, such as high-resolution inputs and P2 connections in feature pyramids, demonstrate exceptional efficacy on this dataset by preserving fine-grained details critical for identifying tiny algae cells. Beyond conventional FPN designs, alternative architectures like BiFPN~\cite{tan2020efficientdet}, PANet~\cite{liu2018path}, and NAS-FPN~\cite{ghiasi2019fpn} offer avenues for further exploration, enhancing feature fusion and optimizing multi-scale representation learning. 

While attention mechanisms (e.g., CBAM, Transformers) remain under-explored in this task, existing implementations demonstrate clear benefits: CBAM suppresses cluttered backgrounds through spatial-channel feature discrimination, while Transformers model long-range dependencies between algae cells in dense configurations. Future research could explore advanced attention variants such as ACmix~\cite{pan2022integration} (blending CNN locality with Transformer global modeling) and Coordinate Attention~\cite{hou2021coordinate} (encoding positional awareness) to further address occlusion challenges, scale variations, and computational efficiency in algal monitoring systems.

Conclusively, the participants' success in the competition can be attributed to their strategic use of a combination of data augmentation techniques, innovative model designs, and the integration of advanced modules. As the field of computer vision continues to evolve, more insights and strategies will undoubtedly contribute to further advancements and breakthroughs in microalgae detection research and applications.

 
\section{Conclusion}

The VisAlgae 2023 challenge aimed to assess the efficacy of object detection algorithms in the realm of microalgal cell detection, merging algae research with computer vision technology. Utilizing a high-throughput microfluidic platform, dynamic video data of microalgal cells were collected under various imaging conditions. The dataset featured six microalgal cell types: \textit{Platymonas, Chlorella, Dunaliella salina, Effrenium, Porphyridium, and Haematococcus}. This paper provides a brief description of the challenge, the dataset, and the top solutions from the participants, holding implications at the intersection of biology and computer vision and representing a step towards harnessing the power of computer vision to unlock new insights into the world of microalgae.

\textbf{Future Work} In future algae object detection competitions, evaluation metrics will not only focus on mAP but also incorporate model size, computational complexity, and inference speed to align more closely with real-world applications. This shift emphasizes practicality and efficiency, requiring solutions that balance accuracy with resource constraints.

\section*{CRediT authorship contribution statement}

\textbf{Mingxuan Sun}: Writing -- original draft, Writing -- review and editing, Data curation, Formal analysis, Visualization.\textbf{Juntao Jiang}: Writing -- original draft, Writing -- review and editing, Conceptualization, Data curation, Formal analysis, Investigation, Methodology, Project administration, Validation, Visualization.\textbf{Zhiqiang Yang}: Methodology, Visualization.\textbf{Shenao Kong}: Methodology, Visualization.\textbf{Jiamin Qi}: Data curation, Visualization.\textbf{Jianru Shang}: Data curation, Visualization.\textbf{Shuangling Luo}: Data curation, Visualization.\textbf{Wanfa Sun}: Methodology, Visualization.\textbf{Tianyi Wang}: Methodology.\textbf{Yanqi Wang}: Methodology.\textbf{Qixuan Wang}: Methodology, Visualization.\textbf{Tingjian Dai}: Methodology.\textbf{Tianxiang Chen}: Methodology, Visualization.\textbf{Jinming Zhang}: Methodology.\textbf{Xuerui Zhang}: Methodology, Visualization.\textbf{Yuepeng He}: Methodology, Visualization.\textbf{Pengcheng Fu}: Writing -- review \& editing, Conceptualization.\textbf{Shizheng Zhou}: Project administration, Data curation, Investigation.\textbf{Qiu Guan}: Methodology.\textbf{Yanbo Yu}: Data curation.\textbf{Qigui Jiang}: Data curation.\textbf{Qiu Guan}: Methodology.\textbf{Liuyong Shi}: Writing -- review \& editing.\textbf{Teng Zhou}: Writing -- review \& editing.\textbf{Hong Yan}: Writing -- review \& editing, Conceptualization, Project administration, Supervision.

\section*{Declaration of competing interest}

 The authors declare that they have no known competing financial interests or personal relationships that could have appeared to influence the work reported in this paper.

 \section*{Acknowledgements}

This research did not receive any specific grant from funding agencies in the public, commercial, or not-for-profit sectors.

 \section*{Data availability}
The dataset is made publicly accessible through the website at https://github.com/juntaoJianggavin/Visalgae2023.

\section*{Declaration of generative AI and AI-assisted technologies in the writing process}

During the preparation of this work the authors used DeepSeek in order to improve language. After using this tool, the authors reviewed and edited the content as needed and take full responsibility for the content of the published article.

\bibliographystyle{cas-model2-names}

\bibliography{references}



\end{document}